\documentclass[]{spie}  

\usepackage{amsmath,amsfonts,amssymb}
\usepackage{subcaption}
\usepackage{graphicx}
\usepackage{cite} 
\usepackage{times}
\usepackage{epsfig}
\usepackage{amsmath}
\usepackage{nccmath}
\usepackage{amssymb}
\usepackage{mwe}
\usepackage{acro}
\usepackage{amssymb}
\usepackage{xcolor,colortbl}
\usepackage{tabularx}
\usepackage{relsize}
\usepackage{pifont}
\usepackage{booktabs} 
\usepackage{multirow}
\usepackage{multicol}
\usepackage{adjustbox}
\usepackage{float}
\usepackage{graphicx}
\usepackage{makecell}
\usepackage{tabu}
\usepackage[colorlinks=true, allcolors=blue]{hyperref}
\usepackage[capitalize]{cleveref}

\usepackage{color}

\colorlet{Mycolor1}{green!10!orange}

\title{Cell Spatial Analysis in Crohn's Disease: Unveiling Local Cell Arrangement Pattern with Graph-based Signatures}

\author[a]{Shunxing Bao}
\author[b]{Sichen Zhu}
\author[c]{Vasantha L Kolachala}
\author[d]{Lucas W. Remedios}
\author[c]{Yeonjoo Hwang}
\author[e]{Yutong Sun}
\author[d]{Ruining Deng}
\author[d]{Can Cui}
\author[k]{Yike Li}
\author[f]{Jia Li}
\author[g]{Joseph T. Roland}
\author[f]{Qi Liu}
\author[h]{Ken S. Lau}
\author[c]{Subra Kugathasan}
\author[b]{Peng Qiu}
\author[i,j]{Keith T. Wilson}
\author[i,j]{Lori A. Coburn}
\author[a,d]{Bennett A. Landman}
\author[a,d]{Yuankai Huo}

\affil[a]{Department of Electrical and Computer Engineering, Vanderbilt University, Nashville, TN, USA}
\affil[b]{Department of Biomedical Engineering, Georgia Institute of Technology and Emory University, Atlanta, GA, USA}
\affil[c]{Division of Pediatric Gastroenterology, Department of Pediatrics and Pediatric Research Institute, Children’s Healthcare of Atlanta, Emory
University School of Medicine, Atlanta, GA, USA}
\affil[d]{Department of Computer Science, Vanderbilt University, Nashville, TN, USA}
\affil[e]{School of Electrical and Computer Engineering, Georgia Institute of Technology, Atlanta, GA, USA}
\affil[f]{Department of Biostatistics, Vanderbilt University Medical Center, Nashville, TN, USA}
\affil[g]{Department of Surgery, Vanderbilt University, Medical Center, Nashville, TN, USA}
\affil[h]{Department of Cell and Developmental Biology, Vanderbilt University School of Medicine, Nashville, TN, USA}
\affil[i]{Division of Gastroenterology, Hepatology, and Nutrition, Department of Medicine, Vanderbilt University Medical Center, Nashville, TN, USA}
\affil[j]{Veterans Affairs Tennessee Valley Healthcare System, Nashville, TN, USA}
\affil[k]{Department of Otolaryngology-Head and Neck Surgery, Vanderbilt University Medical Center, Nashville, TN, USA}

\begin{document} 
\maketitle

\begin{abstract}
Crohn's disease (CD) is a chronic and relapsing inflammatory condition that affects segments of the gastrointestinal tract. CD activity is determined by histological findings, particularly the density of neutrophils observed on Hematoxylin and Eosin stains (H\&E) imaging. However, understanding the broader morphometry and local cell arrangement beyond cell counting and tissue morphology remains challenging. To address this, we characterize six distinct cell types from H\&E images and develop a novel approach for the local spatial signature of each cell. Specifically, we create a 10-cell neighborhood matrix, representing neighboring cell arrangements for each individual cell. Utilizing t-SNE for non-linear spatial projection in scatter-plot and Kernel Density Estimation contour-plot formats, our study examines patterns of differences in the cellular environment associated with the odds ratio of spatial patterns between active CD and control groups. This analysis is based on data collected at the two research institutes. The findings reveal heterogeneous nearest-neighbor patterns, signifying distinct tendencies of cell clustering, with a particular focus on the rectum region. These variations underscore the impact of data heterogeneity on cell spatial arrangements in CD patients. Moreover, the spatial distribution disparities between the two research sites highlight the significance of collaborative efforts among healthcare organizations. All research analysis pipeline tools are available at \url{https://github.com/MASILab/cellNN}.

\end{abstract}



  


\section{INTRODUCTION}
\label{sec:intro}  
Crohn’s disease (CD) is a complex inflammatory bowel disease (IBD) affecting the gastrointestinal tract, characterized by persistent and recurring bowel inflammation ~\cite{baumgart2012crohn}. The prevalence of IBD has been on the rise, leading to increased medical expenditures. Notably, the medical cost of CD was estimated to be \$3.48 billion per year in 2015, and it is projected to reach \$3.72 billion per year by 2025, constituting a significant portion of the overall US national costs ~\cite{hamdeh2020early}. The Gut Cell Atlas Crohn’s Disease Consortium is an ambitious initiative supported by The Leona M. and Harry B. Helmsley Charitable Trust, aiming to develop comprehensive cellular reference maps for CD. The primary focus of this initiative is to compare tissues from CD patients in comparison to healthy controls (\url{https://www.gutcellatlas.helmsleytrust.org/}). By mapping different human cell types and analyzing gene and protein expression in the context of response to anatomical locations and CD, this project offers a unique opportunity to advance our understanding of the human gut and its implications in CD.


In tumor research, spatial analysis using H\&E-stained images has been extensively explored. For instance, Failmezger et al. revealed key tumor microenvironment features through topological tumor graphs in melanoma specimens ~\cite{failmezger2020topological}, while Xu et al. assessed tumor mutational burden and immune infiltrates in bladder cancer patients using H\&E and iHC imaging ~\cite{xu2021machine}. They also developed a method to detect prognostic tumor infiltrating lymphocytes (TIL) density in colorectal carcinoma patients ~\cite{xu2022spatial}. Saltz et al. generated TIL maps from H\&E images, correlating them with survival in diverse tumor types ~\cite{saltz2018spatial}. However, applying spatial analysis to inflammatory bowel disease (IBD) remains under investigation.

\begin{figure*}[t]
\begin{center}
\includegraphics[width=0.8\linewidth]{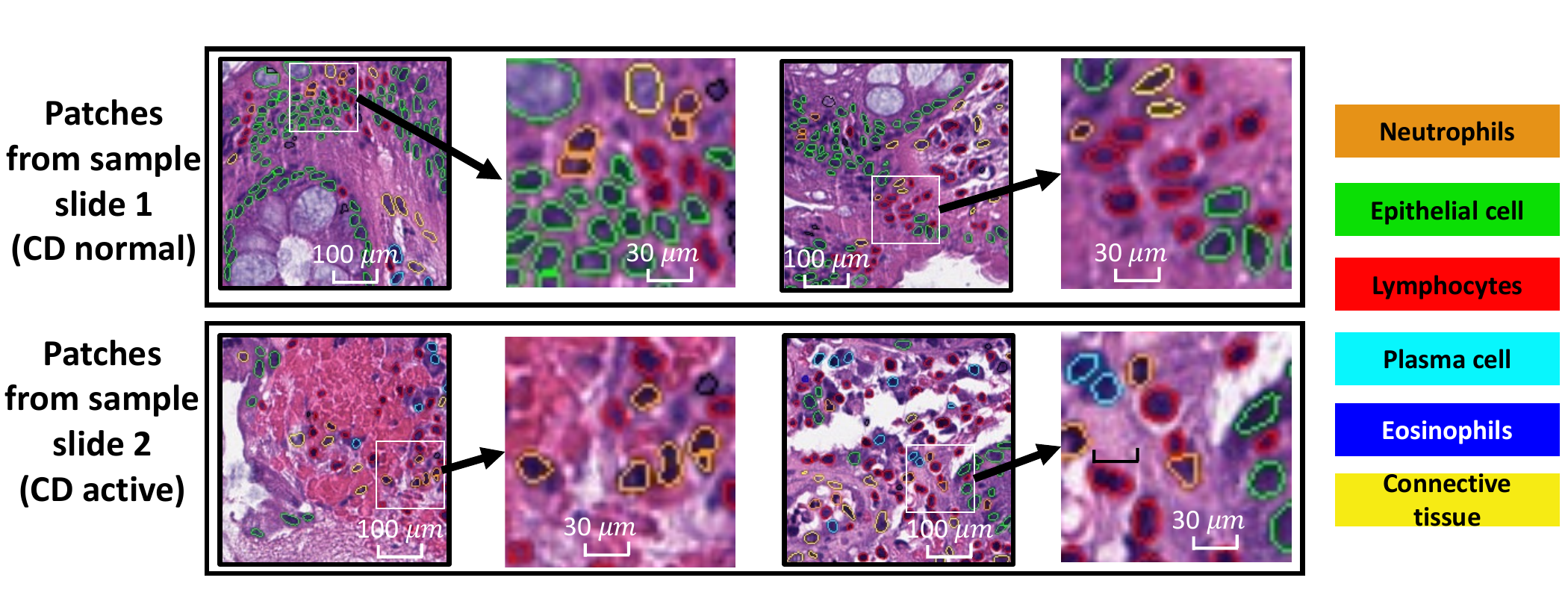}
\end{center}
\caption{We present four patches collected from two sample slides, where one slide is CD normal, and the other slide is diagnosed as CD active. Neutrophils can be found in both types of the tissues. Counting the density of neutrophils is one of the pivotal biomarkers used in identify CD activity. Additionally, pathologists have the capability to zoom in and observe the morphology of individual nuclei or clusters. Our objective is to explore and quantify a more comprehensive morphometric pattern within the localized cellular arrangement.}
\label{fig:problem}
\end{figure*}

Determination of CD activity depends on histological findings, particularly the density of neutrophils observed via hematoxylin and eosin (H\&E) staining ~\cite{pai2022measuring}. There are many ways to assess the samples from CD patients. For instance, pathologists can understand the cell neighborhood changes with a zoom in and out on the biopsies. The invasion of neutrophils is known to be a sign of active inflammation and is a well-known problem ~\cite{gao2015neutrophil}. Pathologist-assigned disease severity scores for CD biopsies are often given at the slide level, though the disease features that resulted in the scoring might not present homogeneously across the slide. As Figure ~\ref{fig:problem} shows, we can see that the high density of neutrophils is not shown in all areas of the tissue on a slide. Can we quantify the relationship of between cell cells, beyond the magnitude of the neutrophils or other cell types is the primary motivation of this work.

In this study, we deviate from conventional morphological feature-based CD activity classification. Instead, our focus is on delving into and defining a graph-based metric aimed at enabling spatial analysis. Our core hypothesis is that distinct patterns might characterize the relationships among various cell types within the context of the CD. Utilizing established tools to identify six specific cell types from H\&E images, we introduce an innovative approach to capture the local spatial characteristics of each cell. In brief, we create a 10-cell neighborhood matrix that outlines the arrangement of neighboring cells for each individual cell. Through the utilization of t-SNE for non-linear spatial projection in both scatter-plot and Kernel Density Estimation (KDE) contour-plot formats, our research delves into disparities in cellular environments, particularly those linked to the odds ratio of spatial patterns between CD active and control groups. The contribution of this study is three-fold: 
\begin{itemize}
    \item We propose a graph-based signature to represent a local arrangement signature for each cell.
    \item We develop a comprehensive visualization workflow to identify the relationship of signature patterns among active CD and normal CD, and healthy control cohorts.
    \item We present investigations from two research institutes with heterogeneous data acquisition, focusing on the rectum.

\end{itemize}

\section{Method}
\begin{figure*}[t]
\begin{center}
\includegraphics[width=1\linewidth]{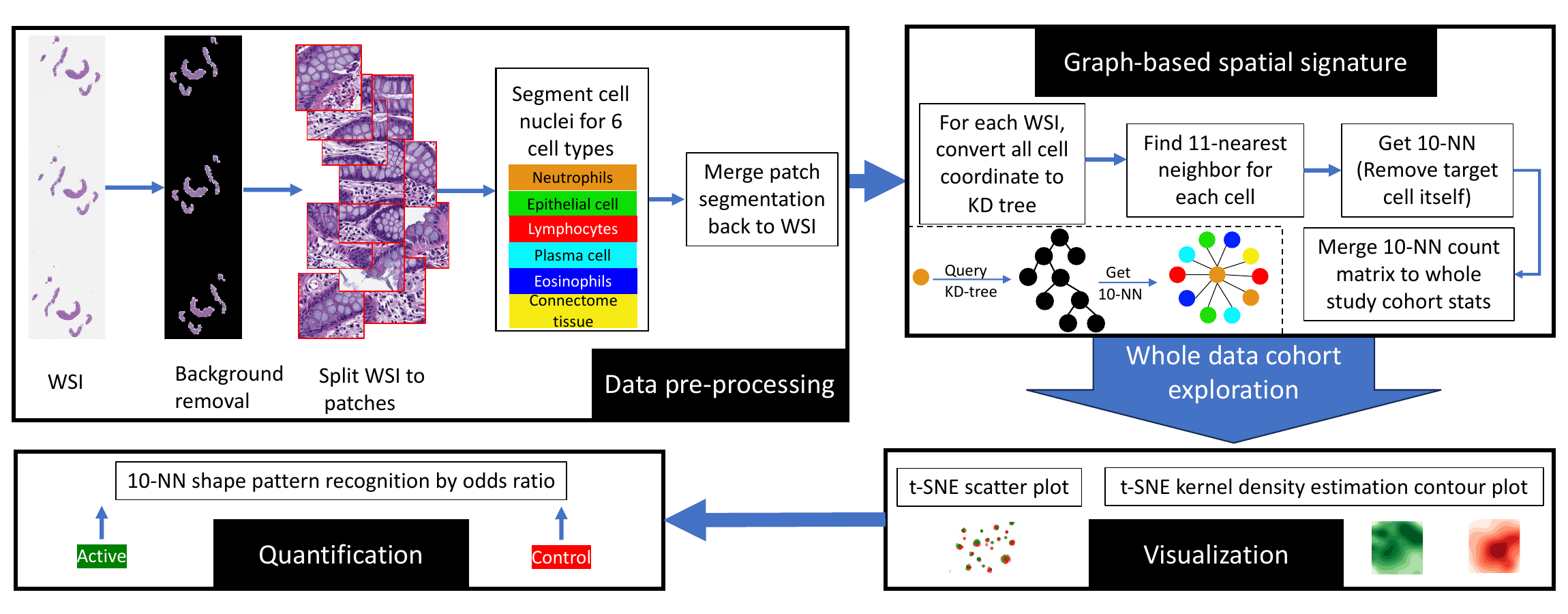}
\end{center}
\caption{This figure illustrates the step-by-step workflow of the whole study data cohort exploration process, showcasing the rigorous methodologies employed in image analysis, cell classification, nearest neighbor calculation, visualization, and ultimate pattern recognition using a graph-based nearest neighbor signature approach.}
\label{fig:workflow}
\end{figure*}

Our objective is to comprehend alterations in the cellular neighborhood. To accomplish this, it is essential to establish biomarkers capable of detecting the nuanced local histological orientation and interrelationships among co-located cells. Here, we investigate a potential avenue for achieving this goal. Initially, we outline a graph-based spatial characterization for each cell. Subsequently, we introduce a visualization technique designed to handle a substantial cell count. Lastly, the quantification method is elucidated. The comprehensive analysis workflow is depicted in Figure ~\ref{fig:workflow}.


\subsection{Data pre-processing}

Prior to generating the spatial signature of each cell, we first segment the whole slide image (WSI) using a pre-trained segmentation deep learning model (HoverNet ~\cite{graham2019hover}) from the Colon Nuclei Identification and Counting (CoNIC) Challenge dataset ~\cite{graham2021conic}, which can identify six cell types: neutrophils, epithelial cells, lymphocytes, plasma cells, eosinophils, and connective tissue. The HoverNet CoNIC pre-trained segmentation model operates only on patches with size of 256$\times$256 pixels under 20$\times$ magnification. So we we employ the CLAM ~\cite{lu2021data} method to remove the background of the WSI and divide the gigapixel image into relevant size of patches, segment each patch (in 20$\times$) using the pre-trained model, and then merge all the patches back into the original WSI space to collect the global coordinate of the cells. 

\subsection{Graph-based spatial signature}

\begin{figure*}[t]
\begin{center}
\includegraphics[width=1\linewidth]{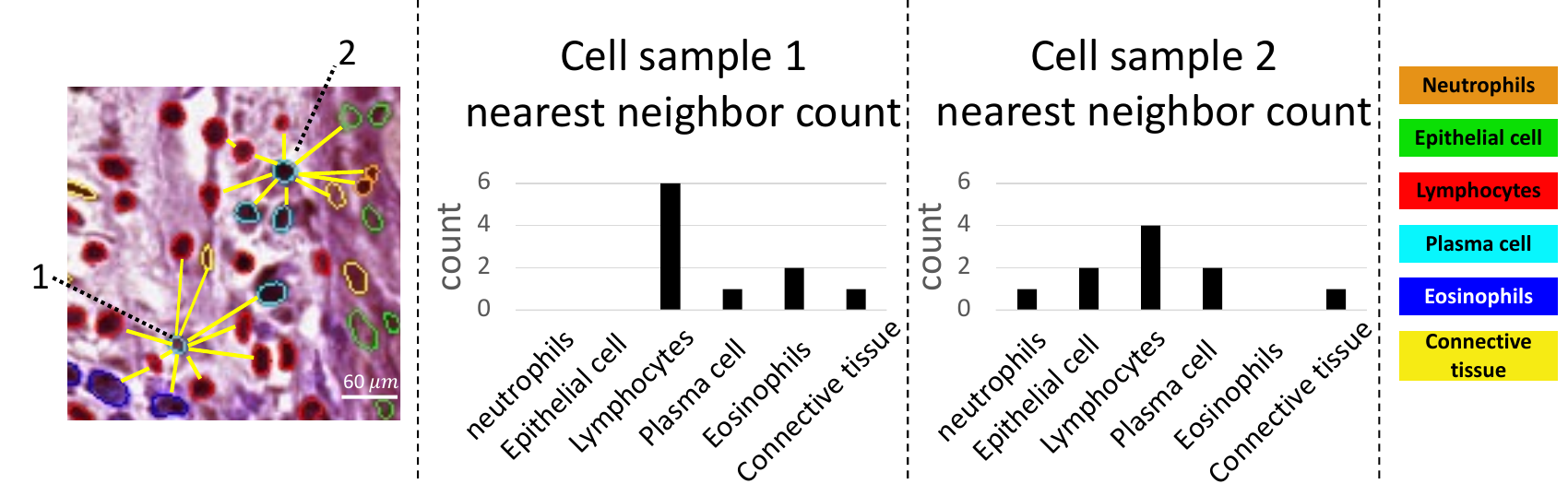}
\end{center}
\caption{Counting the density of neutrophils is one of the key biomarkers to identify CD, which is a well-known problem. We would like to understand if there is a broader morphometry existing within cell local arrangement.}
\label{fig:idea}
\end{figure*}

For building the spatial signature, we aim to find 11-nearest neighbor for each cell, converting the 11 neighbors into a count matrix for the six cell types. Figure \ref{fig:idea} depicts two cell samples of the same type but with different local arrangements. To achieve this, we utilize the KD-tree algorithm ~\cite{pedregosa2011scikit,de2000computational}, which creates a binary tree partitioning the cell coordinate space into smaller regions for fast searching of nearest neighbor points in multi-dimensional spaces. Each cell is assigned with 11 indexes, including itself, resulting in a total count of 10 when excluding the cell itself. To clean up feature noise, we remove edge cases of cells containing less than 10 exclusive neighbors (10-NN).

\subsection{Visualization} 
We aggregate the count matrix of 10-NN signatures for all cells in the target dataset, comprising CD active and control groups, into a large matrix, considering that each WSI may contain over ten thousand cells. Subsequently, we utilize a 2-D standard t-SNE with the KL divergence as the cost function ~\cite{van2008visualizing}. The t-SNE technique effectively maintains the relative distances between neighboring data points, accentuating local structure over global structure. Initially, when employing t-SNE scatter plots, it becomes straightforward to visualize the distinct exclusive regions for the two categories. However, due to the vast number of data points, numerous overlaps are expected, making it difficult to convey the data point density through the scatter plot. As a result, we opt to transform the t-SNE embedding into KDE contour plots to identify regions of interest with diverse probabilities. This visualization approach establishes a non-linear spatial space encompassing all cell data points from the input whole dataset.

\subsection{Quantification} 

We investigate any regions of interest (ROI) of the t-SNE visualization using bounding box (BBox). To comprehend the occurrence probability of the specific 10-nearest neighbor (10-NN) pattern in two different CD activity groups, we calculate the odds ratio. The odds ratio is defined as the fraction of cells from CD activity group 1 divided by the fraction of cells from CD activity group 2 within the BBox.



\section{Experiments and Results}

\subsection{Dataset}
CD can manifest anywhere in the gastrointestinal tract. In our study, we applied our proposed workflow to two datasets acquired from different institutions, with a specific focus on the anatomical region of the rectum. The first dataset was obtained from the Emory University School of Medicine (Emory), also in the Gut Cell Atlas Crohn's Disease Consortium, and includes 8 biopsies from children. Among these, 4 biopsies are from a healthy control group, while the remaining 4 biopsies are classified as CD active.

The second dataset comprises 143 biopsies stained with H\&E, obtained in a deidentified form from Vanderbilt University Medical Center (VUMC) under Institutional Review Board (IRB) approval, specifically Vanderbilt IRB \#191738 and \#191777 ~\cite{bao2021cross}. All biopsies were collected from adult CD patients and scored by a single pathologist, resulting in 97 biopsies classified as normal and 46 biopsies marked as active, categorized into subcategories of mild, moderate, and severe.  It is worth noting that the VUMC dataset lacks a healthy control group. Thus, for a ``normal'' comparator group, the term ``CD normal" refers to patients diagnosed with CD, but the pathologic review of collected tissues were normal, and showed no acute or chronic changes on pathology due to medical therapies.

\subsection{Experiment design}
For each cell type category, which includes neutrophils, epithelial cells, lymphocytes, plasma cells, eosinophils, and connective tissue, we conduct t-SNE visualization and compute the odds ratios for the research institutes separately. Table ~\ref{tab:data_summary} presents a summary of the data points for each site and the six cell subtype categories. To explore the potential ROIs, we investigate them in both the t-SNE scatter plots, where we identify exclusive regions using BBox, and the t-SNE KDE contour plots, where our attention is on areas with relatively high probabilities. All experiments were executed on a workstation boasting 96 CPU cores, 250 GiB of RAM, and an NVIDIA RTX A6000 GPU.

\begin{table}[]
\centering

\caption{Data summary and cell nuclei type counts across different institutions and patient groups.}
\label{tab:data_summary}

\begin{tabular}{lccccc}
                                                     &                                              & \multicolumn{2}{c}{\textbf{Institute 1: VUMC}}                                    & \multicolumn{2}{c}{\textbf{Institute 2: Emory}}                                         \\ \cline{3-6} 
                                                     & \multicolumn{1}{c|}{\textbf{}}               & \multicolumn{1}{c|}{\textbf{CD normal}} & \multicolumn{1}{c|}{\textbf{CD active}} & \multicolumn{1}{c|}{\textbf{Healthy control}} & \multicolumn{1}{c|}{\textbf{CD active}} \\ \cline{2-6} 
\multicolumn{1}{l|}{\multirow{7}{*}{\textbf{Count}}} & \multicolumn{1}{c|}{Sample biopsies}         & \multicolumn{1}{c|}{97}                 & \multicolumn{1}{c|}{46}                 & \multicolumn{1}{c|}{4}                        & \multicolumn{1}{c|}{4}                  \\ \cline{2-2}
\multicolumn{1}{l|}{}                                & \multicolumn{1}{c|}{Neutrophils (neu)}       & \multicolumn{1}{c|}{16,424}             & \multicolumn{1}{c|}{33,103}             & \multicolumn{1}{c|}{384}                      & \multicolumn{1}{c|}{1,022}              \\ \cline{2-2}
\multicolumn{1}{l|}{}                                & \multicolumn{1}{c|}{Epithelial cell (epi)}   & \multicolumn{1}{c|}{5,528,718}          & \multicolumn{1}{c|}{2,409,172}          & \multicolumn{1}{c|}{50,479}                   & \multicolumn{1}{c|}{209,950}            \\ \cline{2-2}
\multicolumn{1}{l|}{}                                & \multicolumn{1}{c|}{Lymphocytes (lym)}       & \multicolumn{1}{c|}{4,037,761}          & \multicolumn{1}{c|}{2,946,595}          & \multicolumn{1}{c|}{37,326}                   & \multicolumn{1}{c|}{379,809}            \\ \cline{2-2}
\multicolumn{1}{l|}{}                                & \multicolumn{1}{c|}{Plasma cell (pla)}       & \multicolumn{1}{c|}{738,568}            & \multicolumn{1}{c|}{607,164}            & \multicolumn{1}{c|}{9,651}                    & \multicolumn{1}{c|}{21,486}             \\ \cline{2-2}
\multicolumn{1}{l|}{}                                & \multicolumn{1}{c|}{Eosinophils (eon)}       & \multicolumn{1}{c|}{50,173}             & \multicolumn{1}{c|}{61,557}             & \multicolumn{1}{c|}{1,148}                    & \multicolumn{1}{c|}{1,437}              \\ \cline{2-2}
\multicolumn{1}{l|}{}                                & \multicolumn{1}{c|}{Connective tissue (con)} & \multicolumn{1}{c|}{1,406,897}          & \multicolumn{1}{c|}{887,472}            & \multicolumn{1}{c|}{17,838}                   & \multicolumn{1}{c|}{60,172}             \\ \cline{2-6} 
\end{tabular}
\end{table}

\subsection{Results}

We provide the outcomes of spatial pattern analysis pertaining to individual cell types across the two institutes as follows.

\begin{figure*}[t]
\begin{center}
\includegraphics[width=1\linewidth]{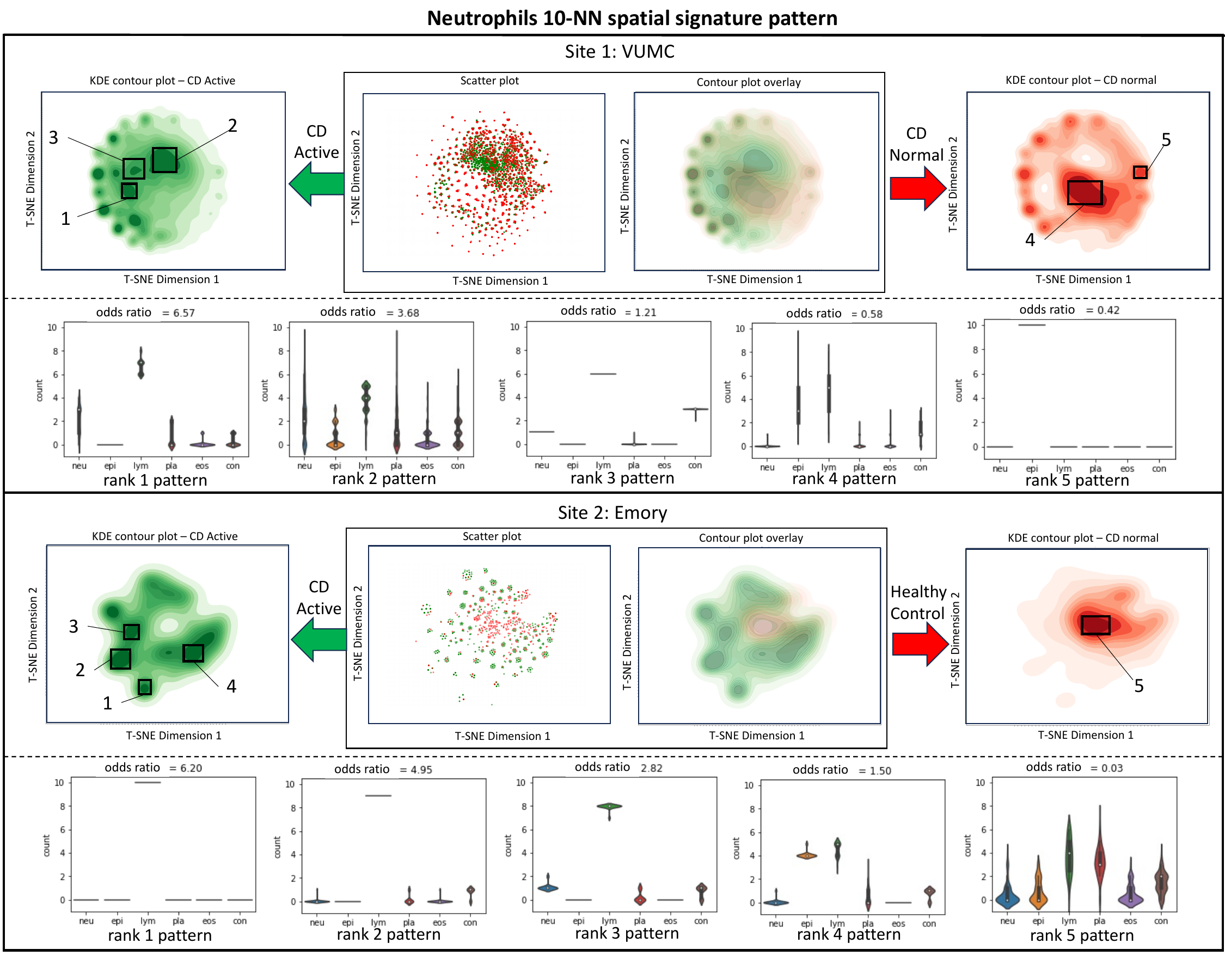}
\end{center}
\caption{Visualization and quantification of neutrophils surrounding a 10-NN shape pattern recognition.}
\label{fig:results_neu}
\end{figure*}

\textbf{Neutrophils} (Figure \ref{fig:results_neu}). The VUMC dataset reveals that CD active tissues tend to exhibit a higher count of 10-NN shapes containing neutrophils and lymphocytes, as well as plasma, with a moderate amount of connective tissue. Moreover, it indicates that when lymphocytes dominate the surroundings without epithelial cells, CD active occurrences surpass those in CD normal tissue. CD normal tissues, on the other hand, tend to have a greater involvement of epithelial cells. In the Emory dataset, it is evident that lymphocytes play a crucial role in the 10-NN interaction with neutrophils. Interestingly, even when both lymphocytes and epithelial cells are dominant, the prevalence remains higher in CD active cases. Furthermore, the dominance of lymphocytes and plasma is associated with a high occurrence rate of healthy controls.

\begin{figure*}[t]
\begin{center}
\includegraphics[width=1\linewidth]{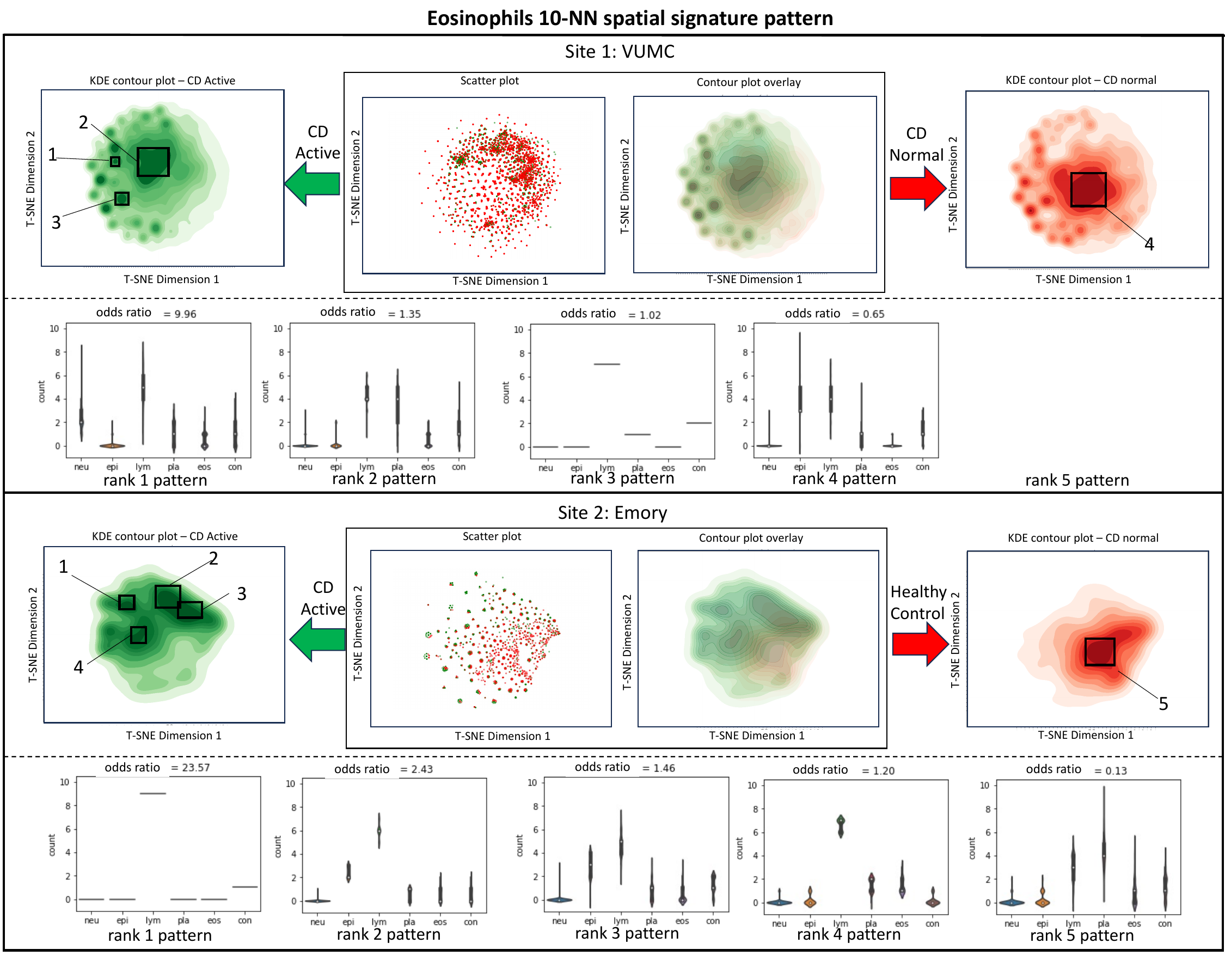}
\end{center}
\caption{Visualization and quantification of eosinophils surrounding a 10-NN shape pattern recognition.}
\label{fig:results_eos}
\end{figure*}

\textbf{Eosinophils (Figure \ref{fig:results_eos})}. In the VUMC dataset, there is a notable significance of lymphocytes and plasma in CD active tissues, whereas eosinophils in CD normal tissues tend to exhibit a higher tendency towards involvement with epithelial and plasma components. However, the Emory dataset reveals a contrasting pattern.

\textbf{Connective (Figure \ref{fig:results_con})}. In the VUMC dataset, shape patterns between CD active and CD normal tissues are remarkably similar. CD active tissues show moderate elevation in lymphocyte and connective tissue presence, while CD normal tissues display increased involvement of epithelial elements. Conversely, the Emory dataset presents greater diversity. It mirrors the VUMC CD active pattern; moreover, dominant connective tissue aligns with healthy controls, and higher plasma presence associates with the control group.

\textbf{Plasma (Figure \ref{fig:results_pla})}. No distinct pattern is identified that is specific to CD active tissues. In the case of CD normal tissues, there is a tendency for an even distribution of involvement among epithelial, lymphocyte, plasma, and connective tissue. This pattern is similarly observed in the Emory dataset. However, when the surroundings solely consist of lymphocytes and plasma, this combination appears more frequently in CD active cases.

\textbf{Plasma (Figure \ref{fig:results_lym})}. In the VUMC dataset, no notable pattern difference in lymphocyte distribution between CD active and CD normal tissues is discerned, where lymphocytes dominate the 10-NN arrangement. This similarity in pattern is also observed in the Emory dataset for CD active cases. However, within the Emory dataset, healthy control cases exhibit a more varied pattern, with the additional presence of plasma and connective tissue in the local environment, alongside lymphocytes.

\textbf{Epithelial (Figure \ref{fig:results_epi})}. Both datasets from the different institutions consistently indicate that epithelial cells predominantly interact with other epithelial cells, regardless of whether the tissue is CD active, CD normal, or from a healthy control.

\section{Discussion and Conclusion}

In this paper, we delve into the exploration of a broader morphometric aspect within the local arrangement of cells, shifting the focus from neutrophil counting to the realm of CD investigation. Our proposed approach introduces a graph-based signature that is customized to portray a distinct local arrangement for each cell. Additionally, we have meticulously designed a comprehensive visualization workflow that illuminates the intricate interconnections among signature patterns within both the CD active patient and CD normal cohorts among adults, as well as between the CD active patient cohort and healthy controls among children. Our study encompasses investigations carried out across two distinct research institutes, centered around the rectum region. The diversified conclusions stemming from shape patterns, extending beyond neutrophils, can be attributed to the diverse array of data acquisition methods. For instance, the VUMC dataset primarily encompasses adult samples and lacks healthy patient tissue, whereas the control group comprises CD biopsies classified as normal. In contrast, the Emory dataset is sourced from pediatric samples, introducing an additional layer of variation in the data collection process. 

In conclusion, our findings indicate that cells do not consistently co-locate in the same manner between CD active and CD normal cohorts for adults rectum dataset, or between CD active and healthy control for children. To delve deeper into these distinct shape patterns, we can leverage RNA-seq to identify cell types, unveil gene expression patterns, facilitate biomarker discovery, enable spatial mapping, and uncover disease mechanisms. Moreover, integrating morphological spatial features, such as considering actual cell distances in the neighborhood of the spatial signature, could potentially enhance our analysis. The observed disparities in spatial distributions across the two research institutes underscore the significance of collaborative endeavors among healthcare institutions.

\section{ACKNOWLEDGMENTS}       
This research was supported by The Leona M. and Harry B. Helmsley Charitable Trust
grant G-1903-03793 and G-2103-05128, NSF CAREER 1452485, NSF 2040462, and in part
using the resources of the Advanced Computing Center for Research and Education (ACCRE) at Vanderbilt University, Nashville, TN. This project was supported in part by the National Center for Research Resources, Grant UL1 RR024975-01, and is now at the National Center for Advancing Translational Sciences, Grant 2 UL1 TR000445-06, the National Institute of Diabetes and Digestive and Kidney Diseases, the Department of Veterans Affairs I01BX004366,I01CX002171, and I01CX002573. The de-identified imaging dataset(s) used for the analysis described were obtained from ImageVU, a research resource supported by the VICTR CTSA award (ULTR000445 from NCATS/NIH), Vanderbilt University Medical Center institutional funding and Patient-Centered Outcomes Research Institute (PCORI; contract CDRN-1306-04869). This work is supported by NIH grants T32GM007347, R01DK135597, and R01DK103831. We extend gratitude to NVIDIA for their support by means of the NVIDIA hardware grant. ChatGPT was utilized for proofreading and grammar checking, and the results were validated by human review to ensure accuracy of facts and intent.

\newpage
\appendix
\label{appendix}

\section{Supplementary information on spatial pattern analysis}

\begin{figure}[H]
\includegraphics[width=1\textwidth]{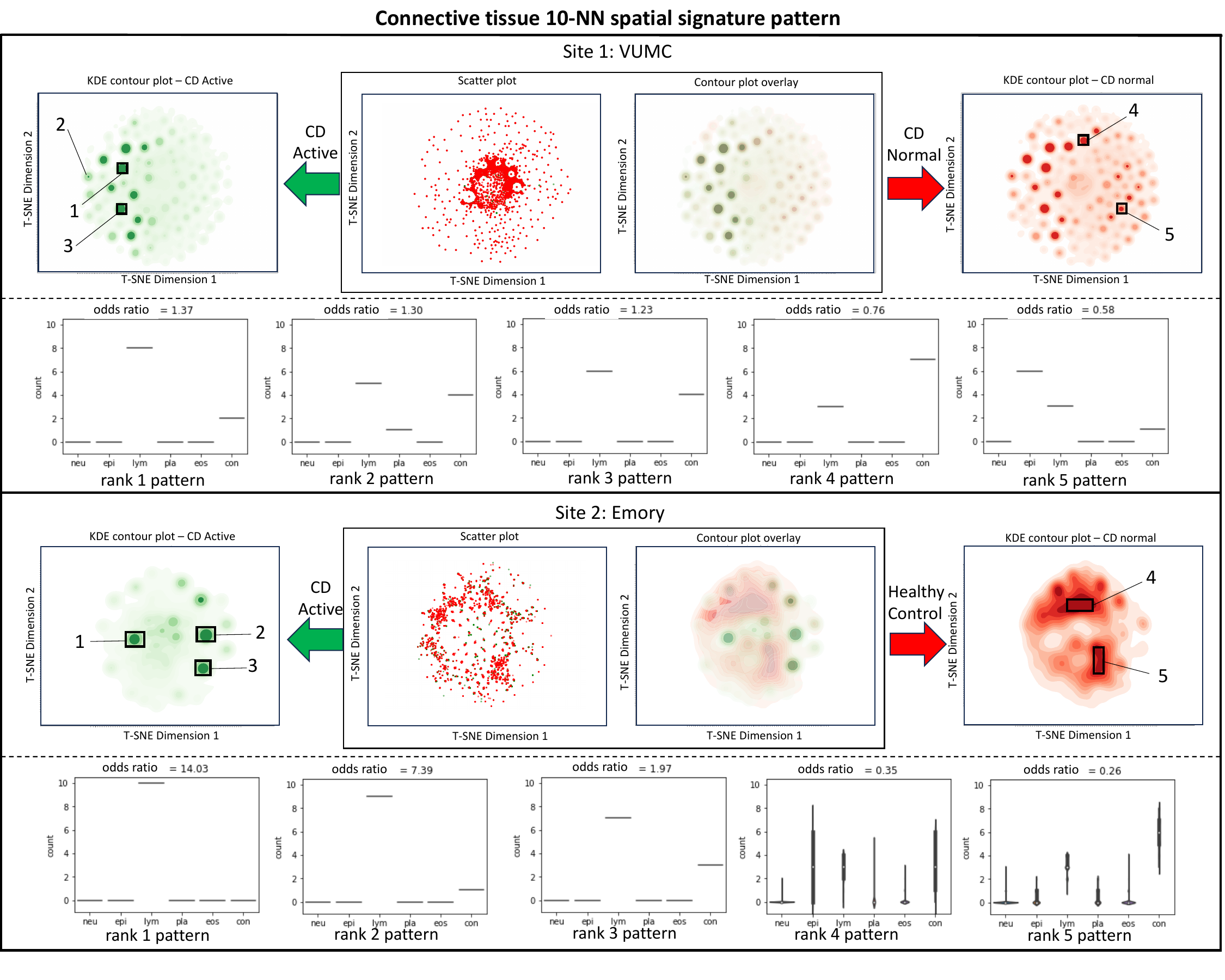}
\centering
\caption{Visualization and quantification of connective tissue surrounding a 10-NN shape pattern recognition.}
\label{fig:results_con}
\end{figure}

\begin{figure}[H]
\includegraphics[width=1\textwidth]{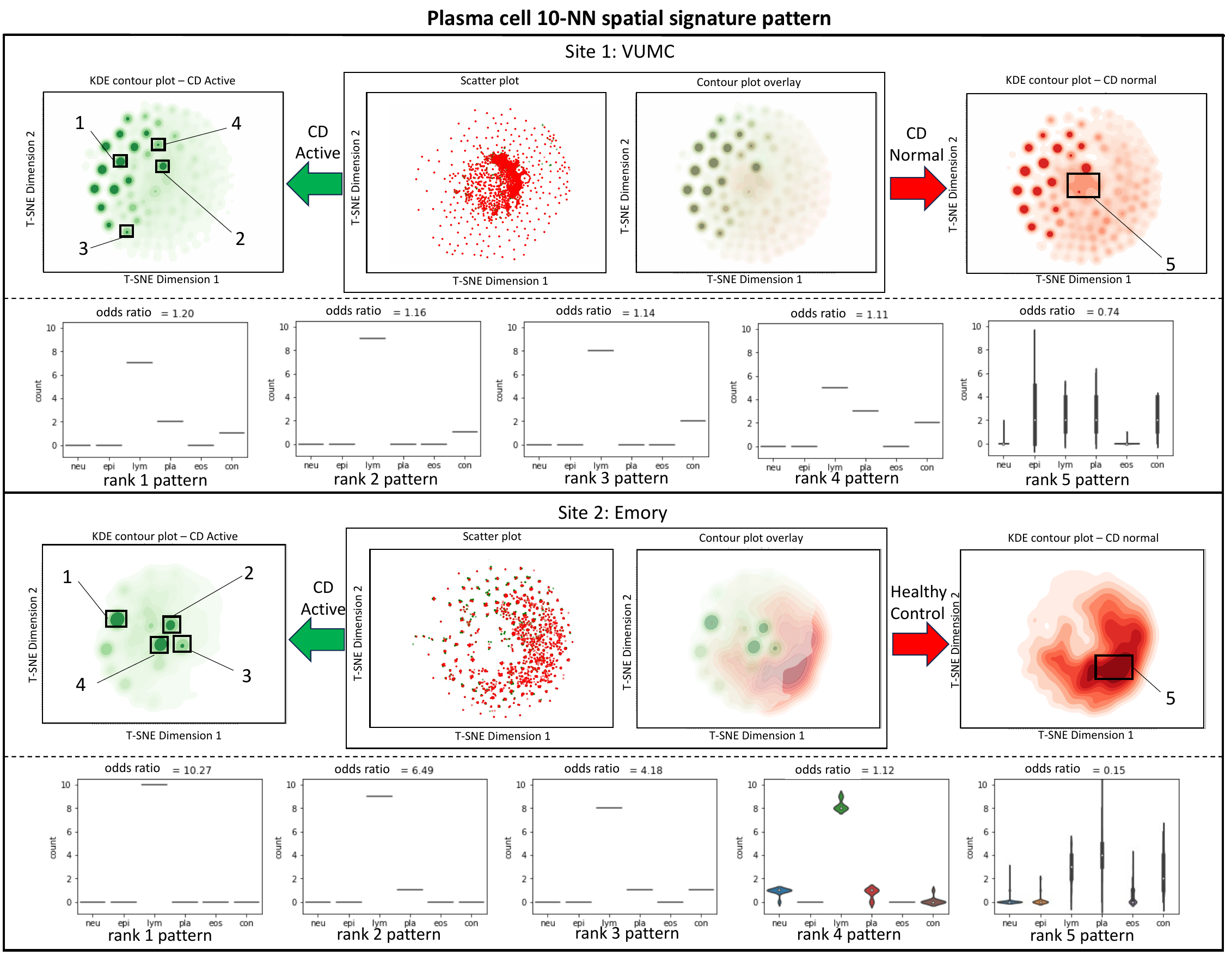}
\centering
\caption{Visualization and quantification of plasma cells surrounding a 10-NN shape pattern recognition.}
\label{fig:results_pla}
\end{figure}

\begin{figure}[H]
\includegraphics[width=1\textwidth]{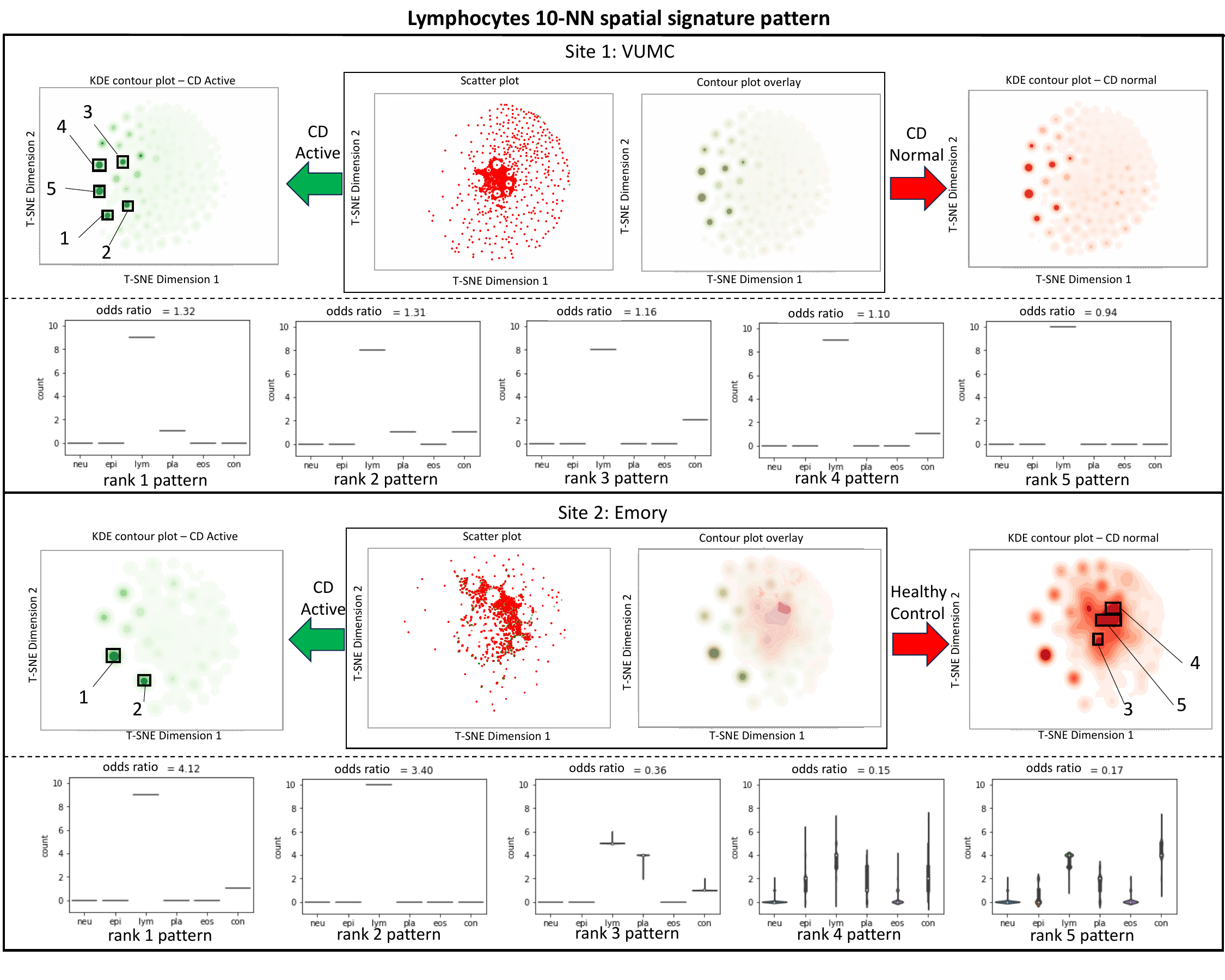}
\centering
\caption{Visualization and quantification of lymphocytes surrounding a 10-NN shape pattern recognition.}
\label{fig:results_lym}
\end{figure}

\begin{figure}[H]
\includegraphics[width=1\textwidth]{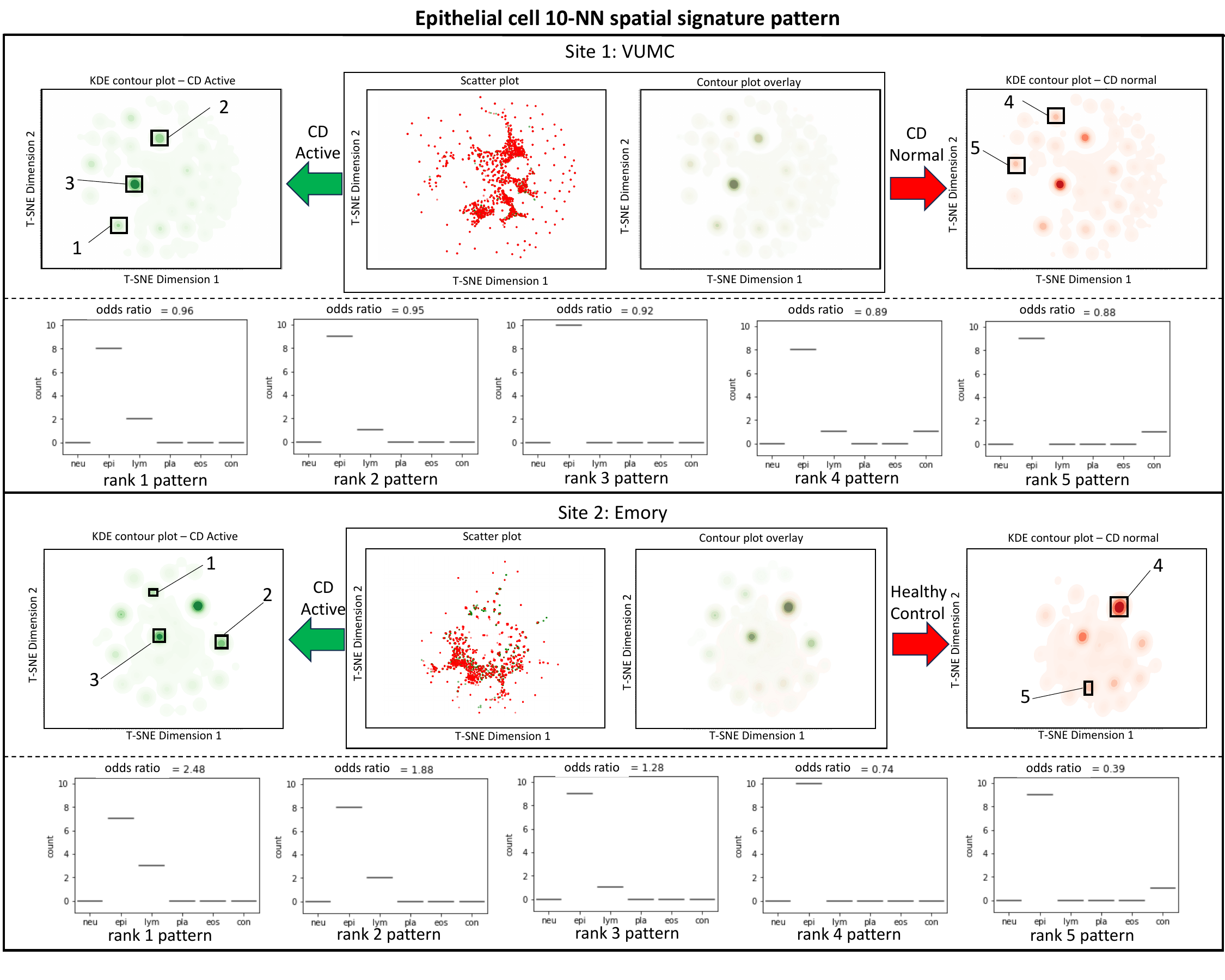}
\centering
\caption{Visualization and quantification of epithelial cells surrounding a 10-NN shape pattern recognition.}
\label{fig:results_epi}
\end{figure}





\bibliography{main} 
\bibliographystyle{spiebib} 

\end{document}